\documentclass[letterpaper]{article}
\usepackage{xcolor}

\usepackage{flairs}
\providecommand{\alt}[1]{}
\usepackage{textcomp} 
\usepackage{graphicx}
\usepackage{xurl}
\usepackage{listings}
\usepackage{amsmath}
\usepackage{amssymb}
\usepackage{float}

\usepackage{times}
\usepackage{helvet}
\usepackage{courier}
\begin{document}
\title{Isolating LLM Lexical Bias: A Curation-Free Triangulated Metric for Preference-Stage Learning}

\author{Xiaoyang Ming, Jose Hernandez, Thomas Stephan Juzek\\
Florida State University\\
tjuzek@fsu.edu\\
}
\maketitle
\begin{abstract}
\begin{quote}
Various language domains have undergone remarkable changes in recent years; these shifts are largely attributed to the advent of Large Language Models and their misalignment with natural language usage. These misalignments are thought to partly originate in the preference-learning stage, e.g.\ Reinforcement Learning from Human Feedback, which generally makes models more useful but simultaneously may introduce systematic lexical bias. In terms of lexical behavior, this is visible in a model's preference for certain formats or the overuse of words (\textit{delve}, \textit{furthermore}), even when such patterns are not present in base model outputs. Research on lexical misalignment induced during preference training is constrained by reliance on manual curation. We address this, by introducing the Triangulated Preference Shift score, a metric that triangulates between human gold standards, base models, and instruct variants to isolate shifts induced specifically by preference learning, without manual curation. We provide data across six model families, anchor the results in the literature, and illustrate the general approach's utility by analyzing whether preference learning shifts models toward what could be interpreted as a ``language of prestige''. The metric provides an initial automated method to quantify behavioral shifts attributable to preference tuning, and thus, may help inform model alignment and development of trustworthy AI.

\end{quote}
\end{abstract}

\section{Introduction}

With the rapid proliferation of AI assistants powered by large language models (LLMs), the deviation behaviors from human expectations in LLM’s responses have arisen as an urgent concern: outputs can be \textit{misaligned} with expectations \cite{saito2023verbosity,sharma2023sycophancy,matsui2024delving,zhang2024lists,twist2025study}. Recent work suggests that preference learning from human feedback (see Section~\ref{sec: preference_background}) can contribute to these shifts, for example by reducing diversity in image generation \cite{zhang2025verbalizedsampling} or inducing conversational and stylistic artifacts such as flattery \cite{bharadwaj2025flattery} and lexical overuse \cite{kobak2024delving,liang2024mapping,liu2024towards,gray2024chatgpt,geng2024chatgpt,juzek2025word}.

\begin{figure}[t]
\centering
\includegraphics[width=\columnwidth]{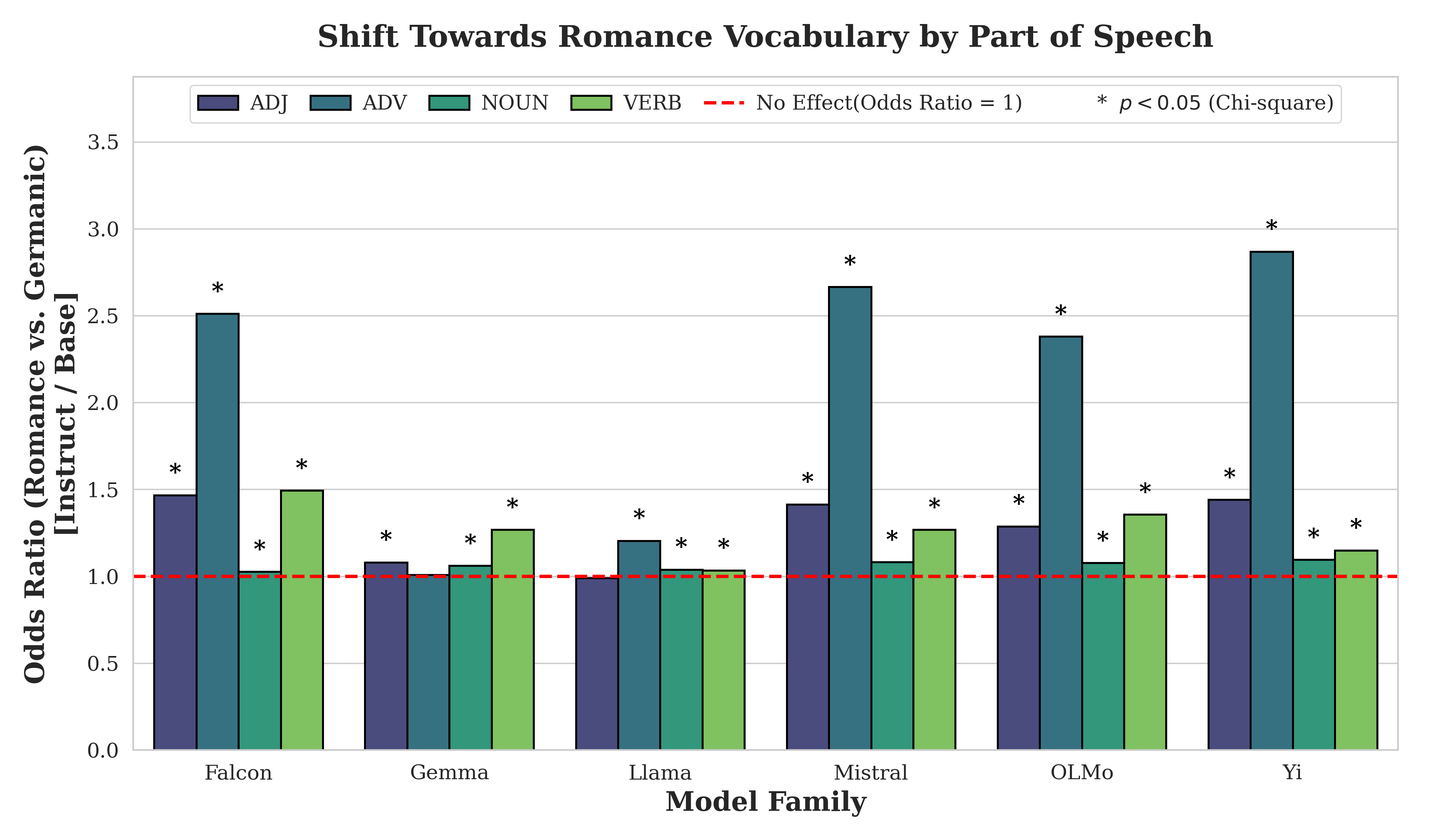}
\alt{Bar chart showing relative shift ratio in the Romance-versus-Germanic content-word ratio between base and instruct model variants across model families.}
\caption{Isolating instruction-stage effects enables linguistic probing of model behavior: instruction tuning shifts instruct models toward prestige-linked vocabulary, increasing the Romance-vs-Germanic content-word ratio relative to base models (significance as per $\chi^2$ test).}
\label{fig:etymology}
\end{figure}

The process of LLM misalignment investigation is typically conceptualized as four major stages: diagnostics (identifying where shifts occur), characterization (describing the nature of the shift), attribution (isolating the specific training stage responsible), and mitigation (correcting or reducing the impact of misaligned behaviors). This article aims to strengthen that diagnostic foundation and to support characterization and attribution as necessary steps toward mitigation. Specifically, to isolate and quantify these preference-associated lexical shifts at scale faces two obstacles. First, identification often relies on manual curation and heuristics. Second, even when behaviors are identified, attributing them to the preference-learning stage is nontrivial.

We address these obstacles by introducing a curation-free pipeline and metric for lexical shift analysis. We obtain generations from six model families, each comprising a base model and an instruction-following variant, using 42,000 PubMed abstracts as prompts, deterministic decoding, and standardized preprocessing (Section~\ref{sec:preprocess}). Then, we compare lexical distributions across human text, base generations, and instruction-following generations, and formalize the result with the Triangulated Preference Shift (TPS) score. TPS quantifies whether overrepresentation is consistent with a shift emerging in instruction-following models toward human usage. We validate TPS in Section~\ref{sec:validation}, illustrate its usefulness on register-linked vocabulary (Section~\ref{sec:etyapp}, also Figure~\ref{fig:etymology}), and discuss its potential in controlled preference-training ablation studies (Section~\ref{sec:ablation}).

\section{Background}

\subsection{Role of Preference Learning in LLM Training Pipeline}
\label{sec: preference_background}
Modern LLM training typically proceeds in the following stages: large-scale pre-training, instruction tuning, preference learning from human judgments, and optional task- or domain-specific fine-tuning \cite{christiano2017drlhf,wei2021finetuned,touvron2023llama2}. Within this flow, preference learning involves presenting the model with multiple candidate responses to the same prompt and training it to favor the output ranked higher by human evaluators \cite{ouyang2022training}. By learning from these relative comparisons, the model is tuned to act with improved helpfulness and safety \cite{ouyang2022training}. Preference learning approaches such as Reinforcement Learning from Human Feedback (RLHF; \citeauthor{ouyang2022training} \citeyear{ouyang2022training}) and Direct Preference Optimization (DPO; \citeauthor{rafailov2024direct} \citeyear{rafailov2024direct}) are widely credited as key drivers of instruction-following performance. Nonetheless, preference learning has been found to be linked to unintended misaligned behavioral shifts in various aspects \cite{zhang2025verbalizedsampling,bharadwaj2025flattery,juzek2025word}. Note that in this article we use the term `preference learning stage' broadly to include the procedures of post-pretraining stages intended to align model behavior with human expectations, in line with the discourse in the literature. 

\subsection{Lexical Misalignment in LLMs}
Among diverse types of misaligned behaviors,  in this article we focus on lexical choices in Scientific English. After ChatGPT's release in 2022, multiple studies reported rapid shifts in Scientific English, with a small set of words (e.g., \emph{intricate}, \emph{nuanced}, \emph{underscore}) increasing sharply over a short time span \cite{matsui2024delving,kobak2024delving,liang2024mapping,liu2024towards,gray2024chatgpt}. However there exist two issues limiting these existing work. First, analyses rely on manual curation and/or post-hoc filtering, which restricts scalability and extension beyond (Scientific) English. Second, the sources of misalignments between human and model lexical behaviors remain under-specified. Thus, further progress requires automated procedures that jointly compare human text, base-model generations, and instruction-following generations so that we can attribute the source of these misalignments to preference learning.

\subsection{Significance of Lexical Misalignment and Societal Impact}
Quantifying these shifts matters. Evidence suggests AI assistance can influence how humans write code \cite{fitterer2025testing} and may even affect spontaneous spoken language \cite{yakura2025empirical,anderson2025model}. More broadly, isolating training-stage contributions connects to concerns about harm from misalignment, including biased outputs, misaligned beliefs, and sycophancy \cite{kotek2023gender,omiye2023large,sharma2023sycophancy,wei2025synthetic,blodgett2020language,bender2021dangers}. Such behaviors raise concerns about normative drift, where repeated exposure to AI-generated language may shift expectations or normalize particular linguistic and ideological patterns \cite{pew2025free_expression,zajonc1968mereexposure,hasher1977frequency}.


\section{Methodologies and Experiments}
\label{sec:meth}

\subsection{Data and Model Families}
\subsubsection{Data}
The literature on lexical overuse focuses on Scientific English. To connect our work to it, our data consists of PubMed abstracts. We randomly sampled 42{,}000 abstracts from 2012--2021 (4{,}200 abstracts per year, without replacement), the decade preceding ChatGPT's release in 2022.

We selected model families based on (i) availability of both base and instruction-tuned variants and (ii) support for deterministic decoding (temperature $T=0$) to ensure reproducibility. 
We selected six model families, specifically utilizing the model with parameter size 7B for Falcon-3 \cite{almazrouei2023falcon}, 4B for Gemma-3 \cite{gemma2025techreport}, 8B for Llama-3.1 \cite{grattafiori2024llama3}, 7B for Mistral \cite{jiang2023mistral7b}, 7B for OLMo-2 \cite{teamolmo2025olmo2}, and 6B for Yi-1.5 \cite{young2024yi}. We used mid-sized variants because larger models often incorporate multimodal capabilities with negligible improvements on the purely textual tasks relevant to this study \cite{huggingface_open_llm_leaderboard_2024}. For model revisions, see our repository (Section~\ref{sec:supp}).

\subsection{Generation, Decoding, (Pre-)Processing}
\label{sec:preprocess}
\subsubsection{Model Generation}
Abstracts were cut into halves at the sentence boundary closest to their midpoints. For each model family, the first half served as the prompt $r_i$ ($i \in \{1, \dots, 42{,}000\}$) to generate both the base-model stream $B$ and the instruct-model stream $I$, while the second half constituted the human gold-standard stream $H$. The approach is related to prior cloze-style work \cite{giulianelli2023midpoint,eisape2020midpoint}. This procedure produced 63.4m tokens across all streams. All generations followed deterministic greedy decoding (Section~\ref{sec:decoding}). After applying the windowing methodology in Section~\ref{sec:windowed-prevalence}, coverage was approximately 2m tokens per model variant.

\subsubsection{Decoding Policy}
\label{sec:decoding}
For reproducibility, we employed deterministic greedy decoding throughout. Sampling was disabled (temperature $T=0$) and nucleus/top-$k$ controls were deactivated. Where randomness applied, we set a global seed. Decoding terminated at the end-of-sequence token \texttt{<eos>} or a maximum length. We enforced a minimum output length of 120 tokens via \texttt{<eos>} suppression and capped generations at 200 tokens. A 4-gram no-repeat constraint mitigated degenerate loops. Base models were treated as pure next-token generators, receiving the prompt $r_i$ as raw input. Instruct models were called with a minimal wrapper consisting of a standardized system message (\texttt{Reply only with the continuation; do not repeat the user text; no preface.}).

\subsubsection{Cleaning and Part-of-Speech Tagging}
\label{sec:cleaning_tagging}
We cleaned all streams in two stages:\ (i) a deterministic regex pass to normalize whitespace and discard text following the abstract's conclusion; (ii) an LLM-based pass using GPT-4o-mini at temperature 0 to remove AI persona markers (``Certainly, here is \dots''), dialogue scaffolding (``\textless assistant\textgreater''), loops (retaining one copy), and first- and second-person material (``Can you explain the meaning of \dots''). The cleaner was instructed to perform deletions only, without paraphrasing or reordering the original text (see full prompt in Section~\ref{sec:supp} repository link ). We applied this procedure to all streams:\ the human gold standard $\{H\}$, the base-model outputs $\{B\}$, and instruct-model outputs $\{I\}$.

To treat inflected forms as single lemmas (``delves'', ``delved'', ..., as the verb ``delve'') and to distinguish homonyms, we part-of-speech (POS) tagged all cleaned texts with spaCy~3.8 \cite{honnibal2020spacy} (model:\ \verb|en_core_web_trf|). Outputs are in the CoNLL--U format \cite{zeman-etal-2017-conll} and contain POS tags from the Universal Dependencies framework \cite{nivre-etal-2020-universal} (UPOS).


\subsection{Evaluation Metric Pipeline}
\label{sec:metric pipeline}
We use the processed data streams $S \in \{H,B,I\}$ to define the Triangulated Preference Shift (TPS) metric, which triangulates the three streams. The metric has two phases, \emph{Estimation} and \emph{Scoring}, built on Windowed Document Prevalence (Section~\ref{sec:windowed-prevalence}). In Estimation, we compute prevalence estimates for all lemma types and store them in a look-up table. In Scoring, we use this table to identify instruct-stage shifts relative to both the base model and the human baseline. This allows us to quantify preference-learning-induced misalignment shifts (Section~\ref{sec:tps}) at both fine-grained lexical resolution and in global comparisons across model families.

\subsubsection{Windowed Document Prevalence}
\label{sec:windowed-prevalence}
Metric reliability depends on the robustness of the lemma-frequency estimator. Simple relative frequency can be unstable, when repeated occurrences concentrate in a small number of documents and skew aggregate statistics. To mitigate this effect and to control for variation in output length across model generations, we adopt a windowed document prevalence approach. For each prompt $r_i$, a percentile offset $\pi_i \in (0,1)$ is chosen deterministically and applied consistently across the $H$--$B$--$I$ triplet. A contiguous \emph{window} of fixed length $K$ tokens is then extracted at this percentile location. If any continuation is too short to support a full window after the symmetric cleaning process (Section~\ref{sec:cleaning_tagging}), the entire triplet is excluded.

To make the procedure concrete, consider the following example (using surface words rather than lemmatized tokens for clarity). Suppose a document consists of:
\begin{quote}
\texttt{This sentence contains five words.}
\end{quote}
This sentence contains 5 words. With $K=2$, there are $5 - 2 + 1 = 4$ possible contiguous windows. If the percentile offset for this prompt is, e.g.,\ $\pi_i = 0.5$, multiplying this value by the number of possible windows gives a starting position index of $0.5 \times 4= 2$. Note that the position index is counted from $0$ at the first word ``This'' (see the index numbers underneath the sentence), so this window contains index 2 and 3 corresponding to word ``contains'' and ``five''. The extracted window is thus:

\begin{quote}
\centering
\small
\begin{tabular}{@{}c@{\hskip 0.8em}c@{\hskip 0.4em}c@{\hskip 0.4em}c@{\hskip 0.8em}c@{}}
\texttt{This} & \texttt{sentence} & \texttt{[contains} & \texttt{five]} & \texttt{words.} \\
\tiny 0 & \tiny 1 & \tiny 2 & \tiny 3 & \tiny 4
\end{tabular}
\end{quote}

\noindent Only the presence or absence of a lemma within this window is recorded. In practice, to ensure a stable estimate of lemma prevalence, we select $M=4$ stratified windows per prompt $r_i$'s document $i$ at different starting indices $\pi$. Let $Y_{iS, m}(w)$ be the binary indicator for lemma $w$ in the $m$-th window of document $i$ for stream $S \in \{H, B, I\}$:
$$Y_{iS, m}(w) = \mathbf{1}\bigl[\, w \in \text{Window}(S, \pi_{i,m}, K) \,\bigr]$$
which takes the value $1$ if lemma $w$ appears at least once within the $m$-th $K$-token window of stream $S$ for prompt $r_i$, and $0$ otherwise. $Y_{iS, m}(w)$ records only the presence or absence of a lemma within $m$-th window, thereby preventing repeated local usage from dominating the signal. 
Then the presence of a lemma in document $i$ is defined as the average across these $M$ windows as $Y_{iS}(w)$, and aggregating $Y_{iS}(w)$ over all $R$ documents in the stream gives the windowed document frequency $c_S(w)$: 
$$Y_{iS}(w) = \frac{1}{M} \sum_{m=1}^{M} Y_{iS, m}(w) \quad c_S(w) = \sum_{i=1}^{R} Y_{iS}(w)$$
Finally, with Jeffreys smoothing \cite{krichevsky1981}, we get lemma $w$ windowed prevalence:
$$l_S(w) = \frac{c_S(w) + \tfrac{1}{2}}{R + 1}.$$

\subsubsection{Metric Definition}
\label{sec:tps}
We penalize shifts already present in the base model and isolate changes emerging in instruction/preference learning. A positive $\mathrm{wTPS}$ occurs only when instruct usage exceeds both the human baseline and the base model, beyond pre-training effects. Using smoothed prevalences, for a model family $M$ we define:
$$\Delta_{IH}(w) = l_I(w) - l_H(w)$$
$$\Delta_{IB}(w) = l_I(w) - l_B(w)$$
$$\Delta_{BH}(w) = l_B(w) - l_H(w)$$

\noindent The lemma-type-level triangulated preference shift is:
\[
\mathrm{wTPS}_M(w) = \min\{\Delta_{IH}(w), \Delta_{IB}(w)\} - \max\{0, \Delta_{BH}(w)\}.
\]
This isolates the ``uplift'' in frequency that is uniquely a product of the preference-learning stage, effectively filtering out misaligned lemmas to identify specific stage-driven drivers of lexical overuse. To score a larger unit $U$ (such as a sequence, document, or corpus), we compute the L2 root-mean-square (RMS) score, denoted as $\mathrm{uTPS}(U; M)$:
\[
\mathrm{uTPS}(U;M) = \left( \frac{1}{|\mathcal{T}(U)|} \sum_{t \in \mathcal{T}(U)} \mathrm{wTPS}_M(\omega(t))^2 \right)^{1/2}.
\]
For clarity, we use $\mathrm{sTPS}_M=\mathrm{uTPS}(s;M)$ for sequence-level scores, $\mathrm{dTPS}_M=\mathrm{uTPS}(d;M)$ for document-level scores, and $\mathrm{cTPS}_M=\mathrm{uTPS}(c;M)$ for corpus-level scores. In the remainder of this paper, we report corpus-level scores separately for base and instruct variants, denoted $\mathrm{cTPS}_{B}$ and $\mathrm{cTPS}_{I}$, respectively.

\section{Results}
\begin{figure}[t]
\centering
\includegraphics[width=\columnwidth]{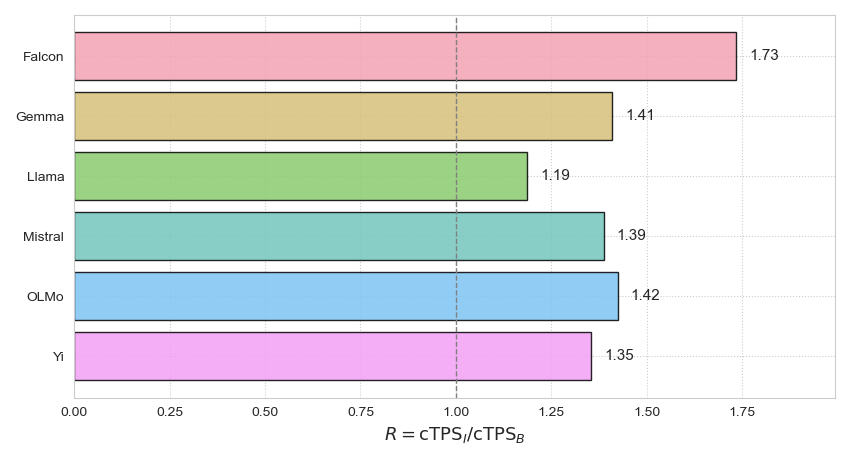}
\alt{Bar chart of corpus-level TPS ratios R = cTPS_instruct / cTPS_base across model families. Higher bars indicate stronger preference-stage shifts in instruct models relative to base models.}
\caption{TPS Corpus-level Relative Ratio for Instruct vs Base}
\label{fig:ctpsratio}
\end{figure}

We apply the full evaluation pipeline and compute $\mathrm{wTPS}(w)$ scores for all lemma+UPOS types. Table~\ref{tab:top20-wtps-simple} lists the top-ranked items. We also compute the corpus-level TPS relative ratio for each model variant, defined as $R=\mathrm{cTPS}_{I}/\mathrm{cTPS}_{B}$. Figure~\ref{fig:ctpsratio} reports $R$ by model family; higher values indicate a stronger preference-stage shift in the instruct variant relative to its base counterpart. 

At the macro level (Figure~\ref{fig:ctpsratio}), all model families exhibit $R>1$, indicating consistently stronger preference-stage shifts in instruct variants than in their base counterparts. The magnitude of the shift varies by family:\ Falcon-3 shows the largest uplift ($R=1.73$), followed by OLMo-2 ($R=1.42$) and Gemma-3 ($R=1.41$), whereas Llama-3.1 shows more moderate effects ($R=1.19$).

\begin{table}[t]
\centering
\small
\setlength{\tabcolsep}{6pt}
\begin{tabular}{r l r l}
\hline
\textbf{Rk} & \textbf{lemma+UPOS}
& \textbf{Rk} & \textbf{lemma+UPOS} \\
\hline
1  & \verb|to_PART|          & 11 & \verb|study_NOUN|     \\
2  & \verb|these_DET|        & 12 & \verb|,_PUNCT|        \\
3  & \verb|this_DET|         & 13 & \verb|highlight_VERB| \\
4  & \verb|to_ADP|           & 14 & \verb|into_ADP|       \\
5  & \verb|furthermore_ADV|  & 15 & \verb|potential_ADJ|  \\
6  & \verb|such_ADJ|         & 16 & \verb|finding_NOUN|   \\
7  & \verb|research_NOUN|    & 17 & \verb|a_DET|          \\
8  & \verb|for_ADP|          & 18 & \verb|as_ADP|         \\
9  & \verb|additionally_ADV| & 19 & \verb|could_AUX|      \\
10 & \verb|further_ADJ|      & 20 & \verb|crucial_ADJ|    \\
\hline
\end{tabular}
\alt{Top 20 lemma+UPOS items ranked by weighted TPS, shown in two columns.}
\caption{Top 20 lemma+UPOS items ranked by $\mathrm{wTPS}(w)$ aggregated across all model families.}
\label{tab:top20-wtps-simple}
\end{table}

\subsection{Validation}
\label{sec:validation}
To assess robustness, we evaluate TPS sensitivity to (i) the window size $K$ and (ii) the random seed used for window placement, holding all other pipeline settings fixed.

\subsubsection{Varying Window Size $K$}
TPS rankings are stable across $K \in \{40,50,60\}$. Top items are unchanged:\ \textit{to} (\textsc{PART}), proximal determiners (\textit{this}, \textit{these}), and the same cluster of content words (e.g., \textit{furthermore}, \textit{such}, \textit{additionally}), as well as comma punctuation.

\subsubsection{Different Seeds}
Rerunning $K=50$ with $seed \in \{43,44,45,46\}$ gives essentially identical lemma-level rankings: top items are unchanged, with only minor swaps lower down.

\section{Application:\ The Etymology of AI-preferred words}
\label{sec:etyapp}

To illustrate the usefulness of our general approach, we test whether base-to-instruct lexical shifts show systematic etymological structure. Using the same processed streams as in Section~\ref{sec:metric pipeline}, we take the list of preferred lexical items and retrieve their etymologies from Wiktionary \cite{wiktionary}. We assign each item to a coarse origin class:\ Germanic, Romance (e.g., French/Latin), Other (e.g., Greek), or Unknown. For simplicity, we analyse only the Germanic--Romance contrast and exclude Other/Unknown; borrowings may be mediated and the categorisation is coarse. We restrict to content words (NOUN, VERB, ADJ, ADV), since function words are largely Germanic and would dominate the signal. Concretely, we reuse windowed document prevalence and aggregate by word class and origin, separately for each model.

Across word classes and model families, with few exceptions, we observe a base-to-instruct shift towards Romance-origin vocabulary (e.g.,\ ``nuanced'', ``multi-faceted'', ``enhance''). French entered English via the ruling class and acquired high socio-economic status \cite{stockwell2001english}; it remains a primary marker of formality \cite{levin1981formality}. A possible interpretation of this pattern could therefore be that it reflects a shift towards a language of \textit{prestige}. Figure~\ref{fig:etymology} summarizes the results and reports $\chi^2$ significance. See detailed computation in Section~\ref{sec:supp} repository link.

Meanwhile, to evaluate the overall etymological shift, we calculated the aggregated Romance ratio for each model by taking the weighted average of the Romance ratio across the four POS categories, using their respective token mass proportion as weights. We observed that while most models' instruction variants showed a consistent shift toward Romance origins compared to their base variants, the Llama instruction variant presented an exception:\ its aggregate Romance ratio actually decreased. This exception stems from a ``compositional effect.'' Although the Romance ratio rose within nouns and adverbs, these categories saw a significant decline in their mass ratio. This shift in POS mass distribution offset the localized Romance increases, resulting in a net decrease in the aggregate ratio. Llama and other model families' etymological results can be seen in Section~\ref{sec:supp} repository link.

\section{Discussion}
\label{sec:ablation}

The items identified by our procedure (Table~\ref{tab:top20-wtps-simple}) closely align with the discourse:\ Of the Top 30 content words (nouns, verbs, adjectives, and adverbs), 22 feature in the literature \cite{matsui2024delving,kobak2024delving,gray2024chatgpt,liang2024mapping,liu2024towards,juzek2025word}. Further, our top list also contains many function words, consistent with findings that AI-generated language exhibits a distinct syntactic style \cite{zamaraeva2025syntaxtheory,munoz-ortiz2024linguisticcontrasting,reinhart2024fullcomparison}. Because $\mathrm{wTPS}(w)$ requires instruct overuse relative to both humans and base models, these results support the view that much AI-associated lexical behavior originates in instruction tuning and preference learning \cite{juzek2025word,bharadwaj2025flattery}. Cross-family differences further align with model-specific ``AI idiolects'' \cite{rudnicka2025can}; also see Section~\ref{sec:supp} repository link.

TPS is relevant to LLM development:\ as a robust measure of preference-induced shift, it can be integrated into pipelines for misalignment ablation and more transparent, trustworthy models. Specifically, as demonstrated by \cite{abdulhai2026how}, linguistic distortions brought by LLM assistant can exert a consistent effect on the semantics of human writing. While the current application of TPS provides a robust observational measure of these biases (exemplified by our etymological analysis), future work will employ controlled studies to establish the causal inference between specific preference-induced lexical shifts and these broader downstream behavioral consequences.

\section{Conclusion}

We introduced the Triangulated Preference Shift score, a curation-free metric that triangulates human text, base-model generations, and instruction-following generations and quantifies shifts induced during instruction-/preference-learning. Using PubMed continuations across six model families, the metric reliably identifies AI-preferred words consistent with observational findings on lexical overuse and is robust to pipeline choices. We illustrated the general approach's usefulness by probing etymological structure in AI-preferred content words, where we observe a tendency toward Romance-origin vocabulary in instruct outputs. 

Our work matters because the metric lets us attribute lexical misalignment to a specific training stage. This makes the diagnostic signal actionable:\ it enables targeted probes and ablations, and thus, provides a pathway toward model improvement and more trustworthy AI. 
In addition, the metric supports linguistic inquiry, as illustrated by our proof-of-concept etymological analysis.

\section{Limitations}
The Triangulated Preference Shift metric is observational:\ it does not, by itself, establish causality or disentangle instruction tuning from preference optimization. Further, the present study is exploratory, and some limitations of the metric will require future refinement. Nonetheless, the metric provides a reproducible target for future controlled ablations that vary preference-training signal and for broader audits of model-specific lexical ``idiolects'' across domains and languages.

Additionally, regarding domain diversity, our analysis is restricted to PubMed abstracts. While this choice allows us to anchor our results in the established literature on scientific lexical overuse, this specific investigation may bias lexical patterns toward Scientific English. 

As for the etymological analysis, it is intended as a supplementary demonstration of our pipeline. Our classification into Germanic versus Romance origins is an oversimplification used to illustrate general trends; in reality, etymological histories are often non-linear and complex. Consequently, while we observe a shift toward Romance-origin vocabulary, one possible interpretation of this pattern is that it reflects a move toward a ``language of prestige'', though further work is needed to substantiate this interpretation.

Finally, from a technical aspect, the cleaning stage in TPS pipeline partially relies on an LLM (GPT-4o-mini) to remove persona markers and dialogue scaffolding. Although we employed deterministic decoding ($T=0$) and strictly instructed the model to perform deletions only (preserving the original lexical choices), the use of an LLM to process LLM outputs could introduce subtle processing artifacts. 

\section{Acknowledgments}
We thank the Florida State University Research Computing Center for computational resources that supported this research and validation. Adam Aleksic provided the idea for the etymological analysis, for which we are grateful. We also thank Gordon Erlebacher and Zina Ward for their intellectual input. Finally, we thank the anonymous reviewers for their constructive feedback.

\section{Ethical Considerations}
The risks presented in our research work are minimal, since it uses public data and open models. A potential ethical concern is the poor working conditions of the human labor behind preference-learning datasets \cite{perrigo2023kenya,perrigo2025sweatshop}. This is an implementation issue, but it warrants scrutiny. 

\section{Supporting Materials}
\label{sec:supp}
All code, configuration files and raw data for this study's pipeline are publicly available at Github repository \url{https://github.com/fsu-nlp/tps-llm-lhf}. Apart from these, the repository includes the following supplemental information: computational setup, TPS ranking results per model familiy and etymological analysis computation.

\bibliographystyle{flairs}
\bibliography{2026_flairs_tps}

\end{document}